\documentclass{article}

\usepackage[numbers]{natbib}

\usepackage[preprint]{neurips_2024}

\usepackage[utf8]{inputenc} 
\usepackage[T1]{fontenc}    
\usepackage{hyperref}       
\usepackage{url}            
\usepackage{booktabs}       
\usepackage{amsfonts}       
\usepackage{nicefrac}       
\usepackage{microtype}      
\usepackage{xcolor}         

\usepackage{graphicx}
\usepackage{subfigure}
\usepackage{color}
\usepackage{adjustbox}
\usepackage{makecell}
\usepackage{amsmath}

\title{Ultra-Fast Adaptive Track Detection Network}

\author{%
  Hai Ni \quad Rui Wang \quad Scarlett Liu \thanks{Corresponding author: Scarlett Liu}\\ 
  School of Traffic and Transportation Engineering, Central South University \\
  Changsha, China 410075 \\
  \texttt{idnihai@csu.edu.cn, ruiwang0618@csu.edu.cn, scarlett.liu@csu.edu.cn} \\
}

\begin{document}

\maketitle

\begin{abstract}
Railway detection is critical for the automation of railway systems. Existing models often prioritize either speed or accuracy, but achieving both remains a challenge. To address the limitations of presetting anchor groups that struggle with varying track proportions from different camera angles, an ultra-fast adaptive track detection network is proposed in this paper. 
This network comprises a backbone network and two specialized branches (\textit{Horizontal Coordinate Locator} and \textit{Perspective Identifier}).
The \textit{Perspective Identifier} selects the suitable anchor group from preset anchor groups, thereby determining the row coordinates of the railway track. Subsequently, the \textit{Horizontal Coordinate Locator} provides row classification results based on multiple preset anchor groups. Then, utilizing the results from the \textit{Perspective Identifier}, it generates the column coordinates of the railway track.
This network is evaluated on multiple datasets, with the lightweight version achieving an F1 score of 98.68\% on the SRail dataset and a detection rate of up to 473 FPS. 
Compared to the SOTA, the proposed model is competitive in both speed and accuracy.
The dataset and code are available at \url{https://github.com/idnihai/UFATD}.

\end{abstract}

\section{Introduction}
\label{sec:introduction}

The 2023 investigative report from the International Union of Railways (UIC) revealed that over 80\% of incidents in 2022 were due to third parties trespassing on railway lines \citep{uicreport2023}. 
The intrusion of a third party onto the railway can lead to prolonged disruptions to railway transportation, potentially resulting in severe casualties and economic losses. Additionally, there is a risk of activities damaging railways and associated facilities, posing significant safety hazards, and jeopardizing overall railway safety.
Traditional static monitoring is inadequate for providing real-time insights. Therefore, there is an urgent need to introduce a methodology for dynamic, real-time railway track monitoring. 
This is essential to accommodate the complex and variable operational environments and meet the demands of all-weather, high-speed operations \citep{catalano2017optical}. The dynamic monitoring method requires the real-time establishment and updating of safety gaps based on the track structure. The track has emerged as a crucial criterion for assessing whether foreign objects have intruded, as emphasized in the work by Zhong \textit{et al.} \citep{zhong2017geometry}. Currently, computer vision-based track detection primarily utilizes two methods: feature-based detection and model-based detection. Feature-based detection excels in accuracy but is susceptible to interference, and its real-time performance is comparatively lower. Conversely, model-based detection exhibits faster speed and robust anti-interference capabilities.

\begin{figure}
  \centering
  \subfigure[Class 0]{\includegraphics[width=0.24\textwidth]{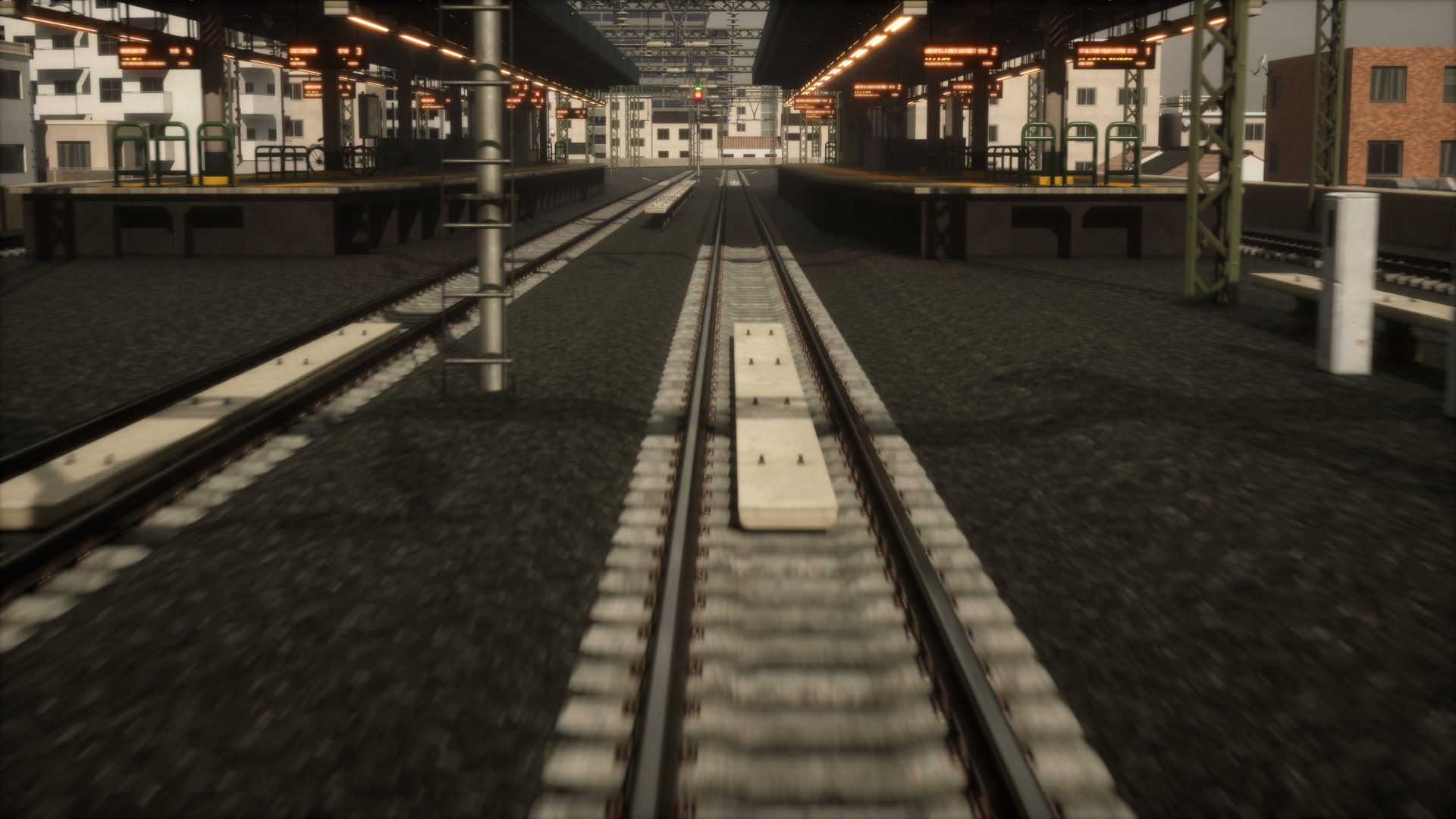}}
  \subfigure[Class 1]{\includegraphics[width=0.24\textwidth]{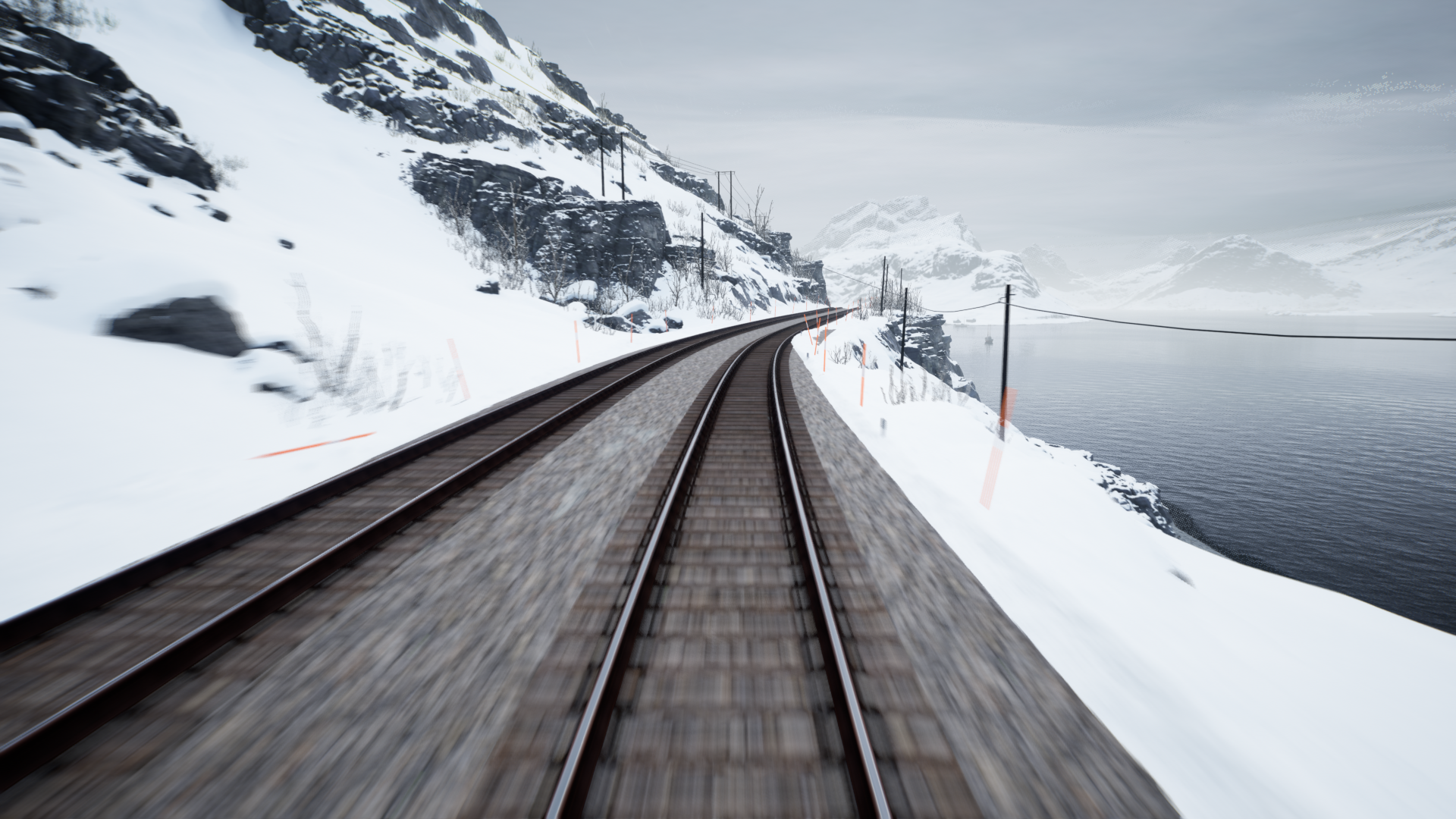}}
  \subfigure[Class 2]{\includegraphics[width=0.24\textwidth]{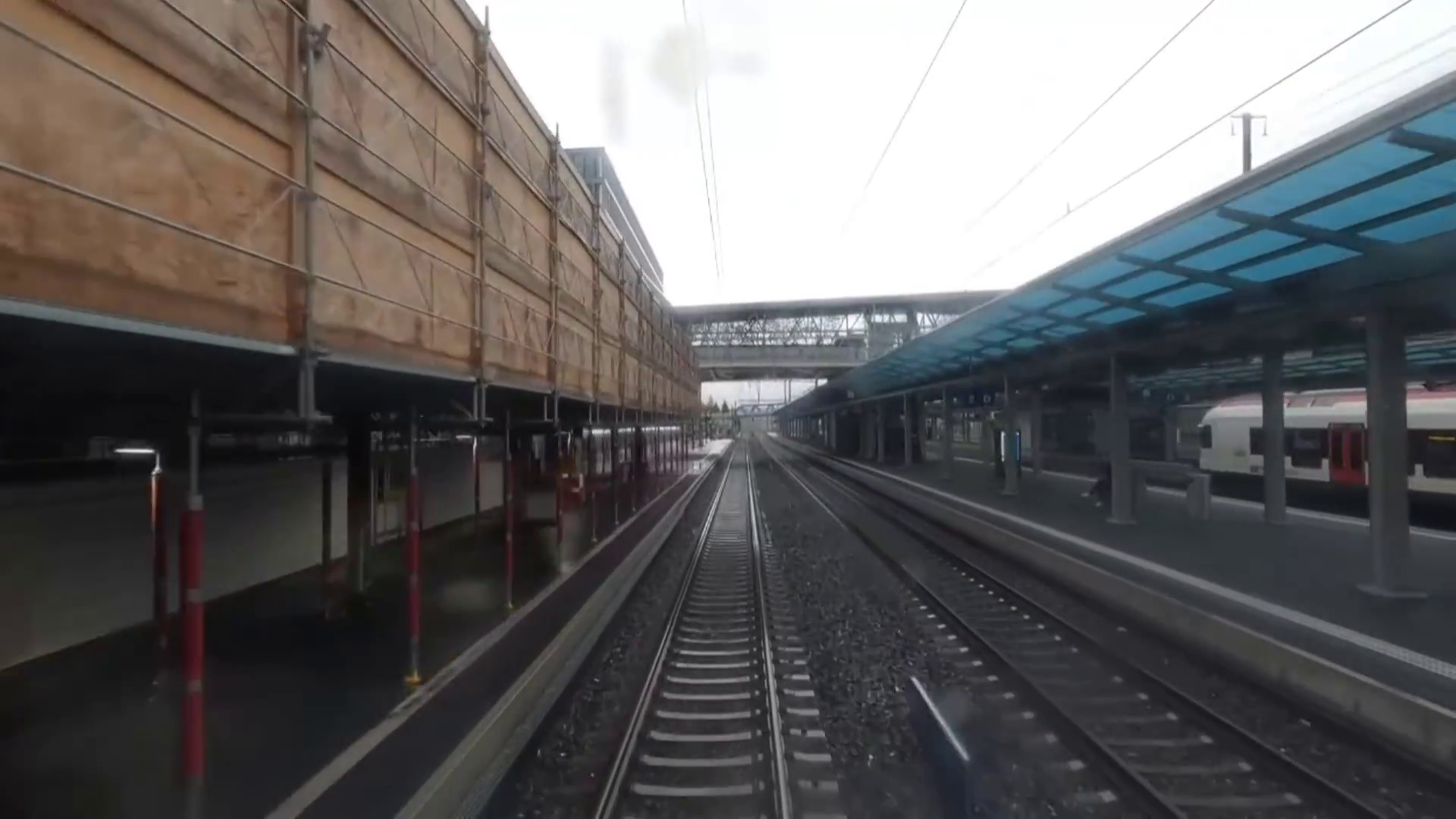}}
  \subfigure[Class 3]{\includegraphics[width=0.24\textwidth]{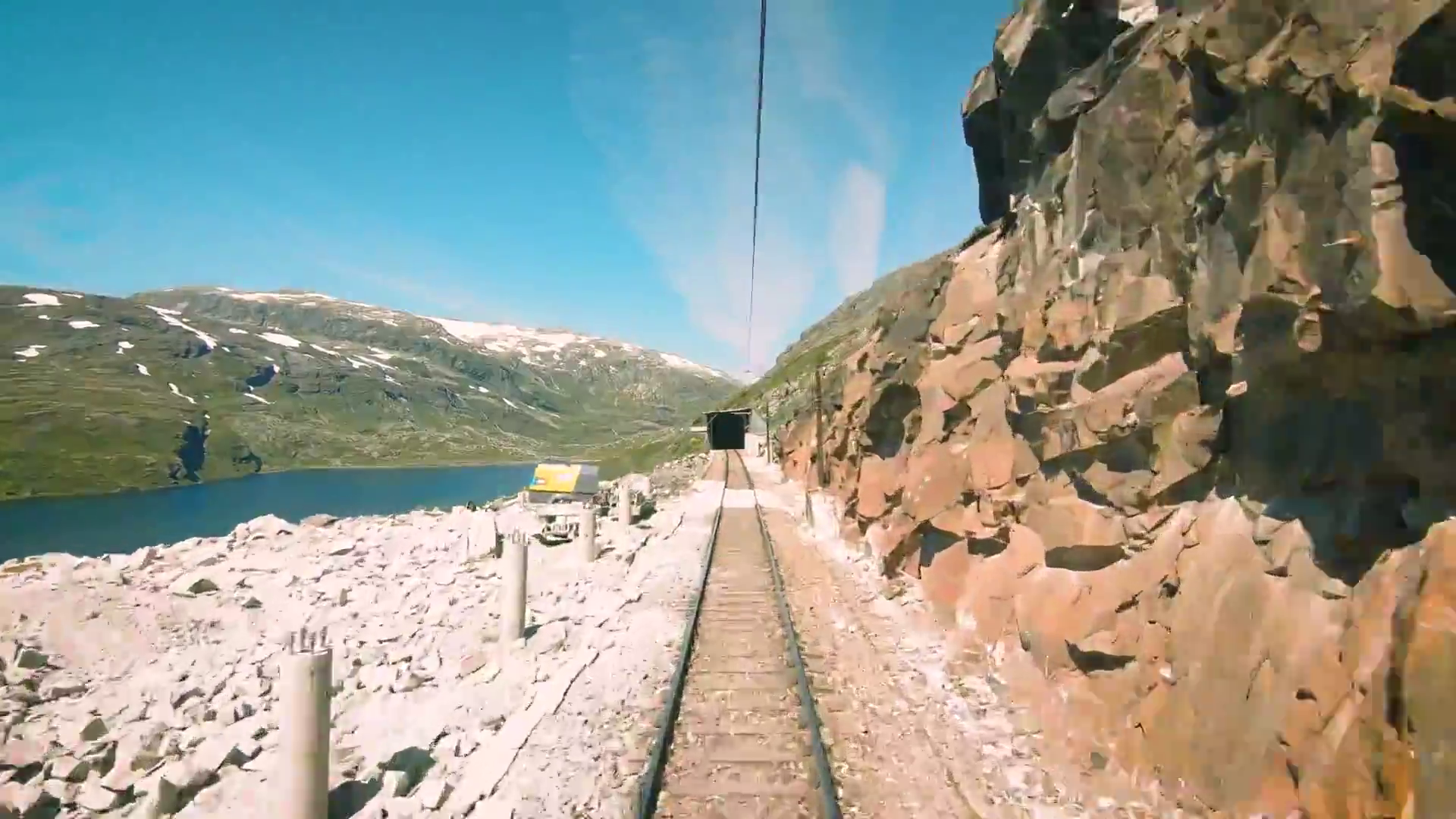}}

  \caption{Railway track images captured from various camera perspectives. As the camera's perspective changes, there is a corresponding alteration in the proportion of sky and ground captured in the images. 
  }
\label{fig:camera_angle}
\end{figure}

The classic model-based method employs image segmentation algorithms \citep{wang2019railnet,giben2015material}, yet it is characterized by high complexity and typically slower speed.
Recently, anchor-based models have been developed \citep{qin2020ultra,li2022rail,qin2022ultra,wang2023lre}.
These methods significantly reduce model complexity and accelerate processing speed through preset anchors.
However, this approach has certain limitations. Typically, anchor-based methods utilize only one group of anchors, which may not be suitable for accurately detecting lanes in scenarios where the lane is choppy.
Significant changes in the angle between the camera and the ground pose challenges, as the preset anchor group may struggle to adequately encompass the entire track line.
Figure \ref{fig:camera_angle}  depicts various instances of disparate camera perspectives, highlighting the challenges a preset group of anchors poses in accurately predicting lane markings.

To accurately detect railway tracks from various angles using UAVs, researchers have proposed a method capable of adaptively representing railways. This method can effectively represent tracks with both near-horizontal and near-vertical inclinations, as well as other tracks with normal inclinations \citep{tong2023trirnet,tong2023anchor}.
Inspired by previous studies, a novel anchor-based row selection method is proposed for railway track detection.
The network classifies the camera perspectives, presets multiple groups of anchors for training, and selects the most suitable groups of anchors corresponding to each camera perspective. This approach can accommodate more intricate scenarios, particularly when the number of anchors is limited.

The contributions are as follows:
\begin{itemize}
    \item A novel anchor-based row selection railway track detection network, UFATD, is proposed, which adapts to different camera perspectives by presetting multiple anchor groups.
    \item Aiming at the situation where UFATD requires multiple groups of anchors, a novel anchor group generation algorithm is proposed.
    \item A synthetic railway track dataset, augmented with a small number of real images, is developed. Its primary purpose is to assess the performance of the novel anchor group generation algorithm across different camera angles.
    
\end{itemize}

\section{Related Work}

\subsection{Feature-based Track Detection}
Feature-based methods initially convert the image to grayscale, followed by the application of edge detection algorithms to extract the tracks.
Some widely employed edge detection algorithms include the Prewitt algorithm \citep{prewitt1970object} and the Sobel algorithm \citep{sobel19683x3}.
Nevertheless, the aforementioned methods do not account for image noise during image processing.
Moreover, to effectively address the constraints caused by railway encroachments, Wang \textit{et al.} devised an adaptive track segmentation algorithm. 
This algorithm enhances detection accuracy by using Gaussian kernels with adaptive orientations, based on the Hough transform's maximum point, to optimally utilize linear features in railway scenes \citep{s19112594}.

Utilizing the Line Segment Detector (LSD) and least squares curve fitting achieves high accuracy in recognizing image tracks. Nevertheless, relying solely on edge features presents challenges, particularly in scenarios featuring physical obstructions or exposure variations \citep{zheng2021rail}.
Employing the Canny operator, which integrates Gaussian theory, structural element function, and morphological transformation, effectively captures track features \citep{zheng2021rail}. This method enhances the algorithm's adaptability and robustness, even in blurry or low-quality videos. Nevertheless, achieving accurate rail detection in environments such as tunnels and shadows remains challenging  \citep{wang2022rail}.

\subsection{Model-based Lane and Track Detection}

When tackling detection challenges in complex scenes, model-based methods typically outperform feature-based methods.
Various models based on deep learning, including image segmentation models \citep{wang2019railnet,giben2015material,laurent2024train,neven2018towards} and anchor-based models \citep{qin2020ultra,li2022rail,qin2022ultra,wang2023lre,tong2023trirnet,tong2023anchor,song2023novel,liu2021condlanenet,yu2024dalnet}, have been developed for lane line detection or railway track detection, providing comprehensive solutions for complex scenes.
The end-to-end segmentation model, TEP-Net, achieved an outstanding IoU of 97.5\% on the RailSem19 dataset \citep{laurent2024train}. Another end-to-end lane segmentation model operates at a speed of only 50 fps \citep{neven2018towards}.

Qin  \textit{et al.} addressed lane detection by regarding it as a row-based selection issue employing global features. They proposed a lane line detection method based on this approach \citep{qin2020ultra}. Li \textit{et al.} innovatively applied the anchor-based method to railway track detection, achieving excellent results. Comparative experiments with segmentation models revealed that Rail-Net excels in both accuracy and speed \citep{li2022rail}. The DALNet model, proposed by Yu \textit{et al.}, incorporates a dynamic anchor line generator, which dynamically creates suitable anchor lines for the railway tracks based on features such as their position and shape in the input image \citep{yu2024dalnet}.
Among these methods, anchor-based methods exhibit superior speed compared to segmentation-based methods. 
Additionally, the detection accuracy is comparable to that of segmentation-based methods.

\subsection{Camera Pose Estimation}

The estimation of the IMU or gyroscope sensor utilized by the robot is impacted by noise from uneven ground, making pose estimation challenging. Drawing inspiration from the classification network rooted in Optical Character Recognition (OCR) \citep{jaderberg2016reading} principles, scholars investigate the application of deep learning for estimating camera image poses to achieve precise camera positioning \citep{kawai2022camera,kawai2023dnn}.
The outcomes of the classification network are utilized to correspond to distinct perspectives.
To accurately represent lane lines, researchers proposed a homography prediction network, termed HP-Net, which predicts the homography matrix of the camera \citep{chen2023improving}.

\section{Data Scope}
\label{sec:datascope}
\subsection{Dataset Collection and Annotation}
\label{sec:data_collect}
To substantiate the efficacy of the proposed method, comprehensive experimental evaluations will be conducted across three datasets, namely: \textit{SRail}, \textit{RailDB} and \textit{DL-RAIL}. The three datasets pertain to railway track imagery. 
The SRail dataset, introduced in this study, comprises 5,903 images. Among these, 4,748 are synthetic images generated using \textit{Unreal Engine 5}\footnote{https://www.unrealengine.com/}, supplemented by 1,155 images captured by real cameras to enhance the dataset's feature diversity.
These images simulate diverse camera angles for a comprehensive assessment.
Images are annotated using \textit{LabelMe}\footnote{https://github.com/wkentaro/labelme}.
 This dataset aims to evaluate the model's performance across various camera perspectives. Accordingly, the images are categorized into four classes based on their camera viewpoint: Class 0, Class 1, Class 2, and Class 3. Sample images from each category are depicted in Figure \ref{fig:camera_angle}.

 The indicator used by Li et al. to evaluate the model on the RailDB dataset is ACC \citep{li2022rail}. 
\(ACC = \frac{\sum_{clip} C_{clip}}{\sum_{clip} S_{clip}}\) where \(C_{clip}\) represents the number of correctly predicted track points, and \(S_{clip}\) denotes the total number of ground-truth data points in every image.
 However, this indicator only evaluates the abscissa (x-coordinate) of the points. The method proposed in this article adjusts the anchors based on the image, resulting in changes to the ordinate (y-coordinate) in the outcome. Therefore, ACC is not suitable for the evaluation of this method. In order to more effectively evaluate the method on the RailDB dataset, the rail image and label image were resized to \(1920 \times 1080\), and txt annotation files were generated based on the label images. In addition, similar to their work, this study focused on only four railway tracks as research objects. Consequently, any values greater than 4 in the label image were set to 0.

\subsection{Data Comparison}

In this section, the differences among the aforementioned datasets are compared. Table \ref{tab:nums_datasets} displays the features of datasets.

\begin{table}[ht]
\centering
\caption{The existing railway dataset. The original RailDB dataset contains images of various sizes. To facilitate the employ of the F1 score evaluation algorithm, all rail images and label images are resized to a uniform size.}
\label{tab:nums_datasets}
\begin{adjustbox}{width=1.\textwidth} 
  \begin{tabular}{ l | c  c c  c }
  
     \toprule
    Database &  Rail-DB & DL-RAIL & SRail (ours) \\ 
   \midrule
    Quantity & 7,432 & 7,004 & 5,903\\ 
    Number of tracks & Original: 8 \quad After processed: 4 &2 &2\\
    Resolution  &  \makecell[c]{Original: 1280 \(\times\) 720;  1920 \(\times\) 1080; 3840 \(\times\) 2160 \\After processed: 1920 \(\times\) 1080 } 
  & 1920 \(\times\) 1080   & 1920 \(\times\) 1080\\
    Train &   5,946 &5435 &4,132 \\
    Test  & 1,486& 1,569 & 1,771\\ 
    \bottomrule

  \end{tabular}
  \end{adjustbox}
\end{table}

\section{Method}
\label{sec:method}

In this paper, track detection is regarded as a row-based selection problem rather than a segmentation problem, which significantly reduces model complexity.

\subsection{Problem Formulation}

The segmentation method typically involves annotating the track lines as multi-valued maps. However, n this study, track detection is regarded as a row-based selection problem and a classification problem. The classification problem aims to identify the anchors corresponding to the image, while the row-based selection problem involves gridding the image based on these anchors. 

Additionally, a fixed number and position of anchors can result in significantly low efficiency in track detection. For instance, if the anchors cover the entire image while the railway track only occupies half of the image, half of the anchors become redundant.
To address this issue, an adaptive method is proposed in this paper.
Specifically, multiple groups of anchors are preset, from which the network dynamically selects the most appropriate group of anchors. This method allows for the representation of the railway tracks with as many anchors as possible.

\begin{figure}[b]
  \centering
  \subfigure[Example 1]{\includegraphics[width=0.49\textwidth]{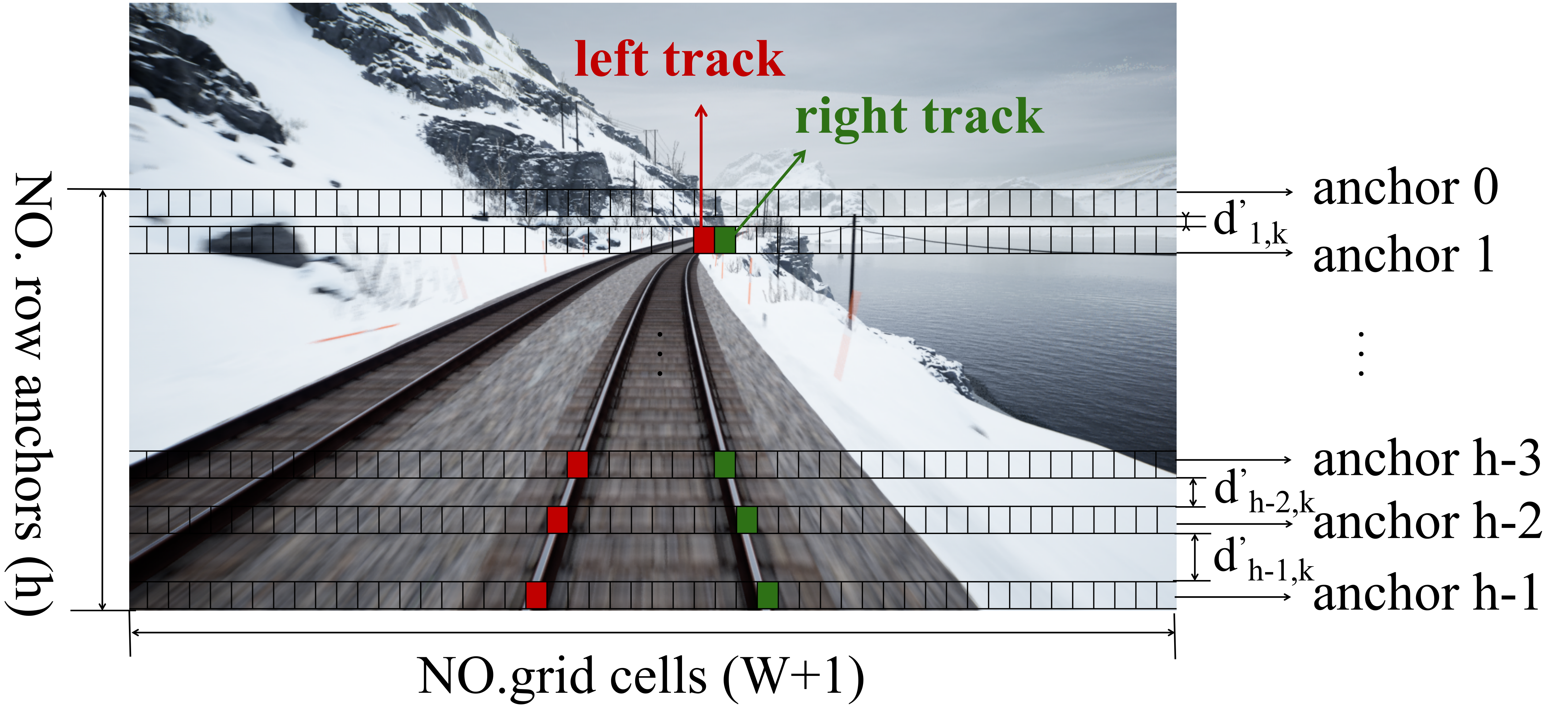}}{\label{fig:select1}}
  \subfigure[Example 2]{\includegraphics[width=0.49\textwidth]{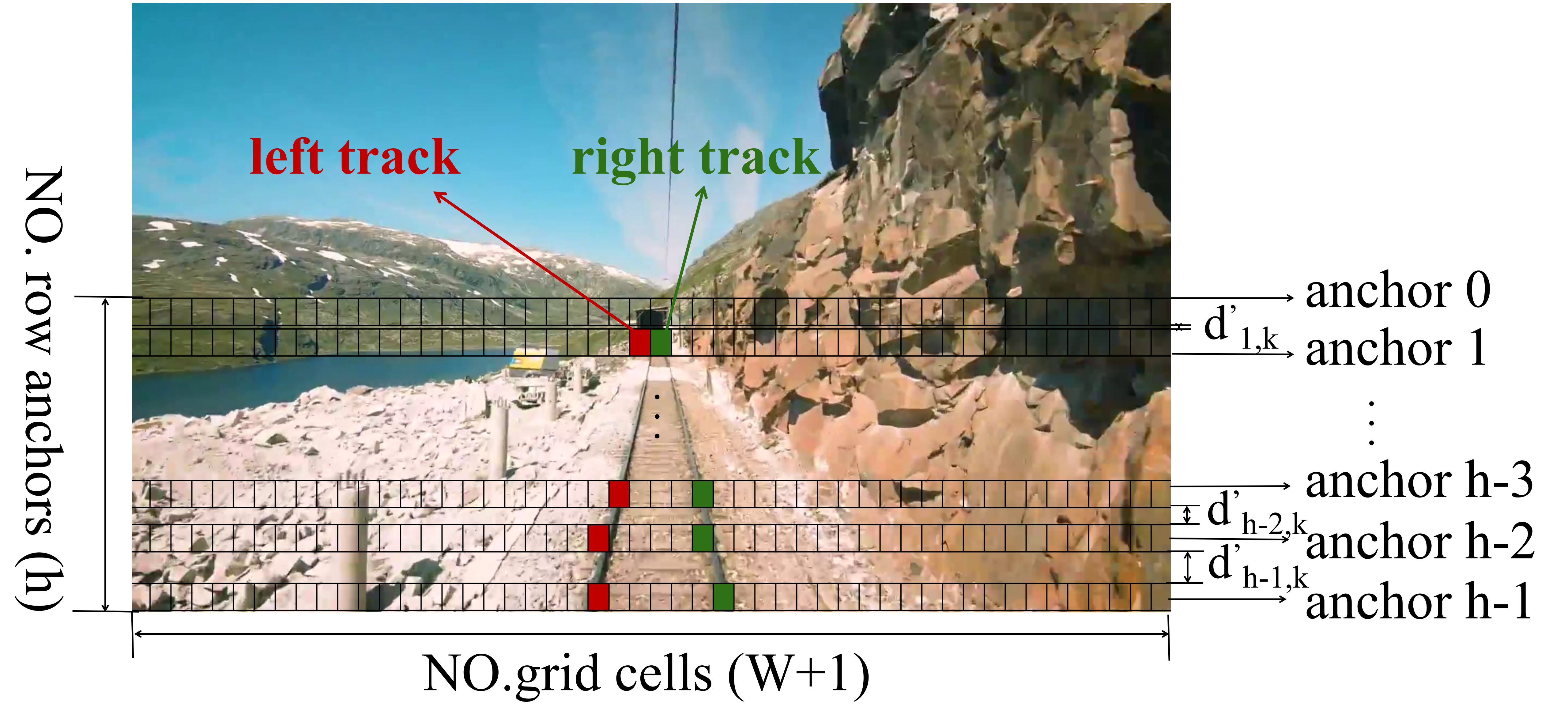}}{\label{fig:select2}}

  \caption{Illustration of gridding the image by selecting a suitable group of anchors for the image. 
  Each image will be matched with an appropriate group of anchors. 
  The image is divided into \((w+1)\)  grids in the horizontal direction based on these anchors.
To better locate possible bends and other features in the upper part of the image, the distance between anchors decreases as they approach the top of the image. 
\(d'_{j,k}\) is the anchor row distance calculated by the method proposed in section \ref{sec:anchor_gen}.
}
\label{fig:anchor_select}
\end{figure}

As illustrated in Figure \ref{fig:anchor_select}, there are n groups of anchors. Within each group, the track line intersects with the row represented by the anchor, thus transferring the labeled polylines of the railway track to a series of horizontal locations and their corresponding anchors.
Different images will correspond to different anchor groups.
The selected anchor groups are designed to cover the track with as many anchors as possible, thereby maximizing the utilization of anchors.
By employing non-uniform anchors, the distribution of anchors at the top of the image becomes denser, yielding improved results when the top of the image contains curved railway tracks.

The computational expenses can be notably diminished using this approach. Assuming \(C\) represents the number of tracks, \(h\) denotes the number of row anchors, and \(w\) signifies the number of grid columns. Analogous to UFLD, \(w\) is expanded to \((w+1)\) in the grid column direction, where \((w+1)\) represents the background. 
Besides, \(n\) represents the number of groups of anchors used for training.
Consequently, the computational cost is proportional to \( ((w+1) \times h\times C\times n)\). When the image size is \(H \times W\), the calculation reduction rate (r) can be expressed as:
\( r = \frac{H \times W}{h  \times (w+1) \times n}\)

In the work of Li Li \textit{et al.} \citep{li2022rail}, the number of anchors \(h\) is set to 52, and \(w\) is set to 200. Simultaneously, the images are also scaled to \(800 \times 288\).
Therefore, the \(r\) can reach up to 22. 
The smaller \(h\) and \(n\) are, the larger \(r\) will be.  When \(h\) is set to 18, and \(n\) is set to 2, \(r\) can reach 31.8.

\subsection{Network Pipeline}
\label{sec:network}

\begin{figure}
  \centering
  {\includegraphics[width=0.88\textwidth]{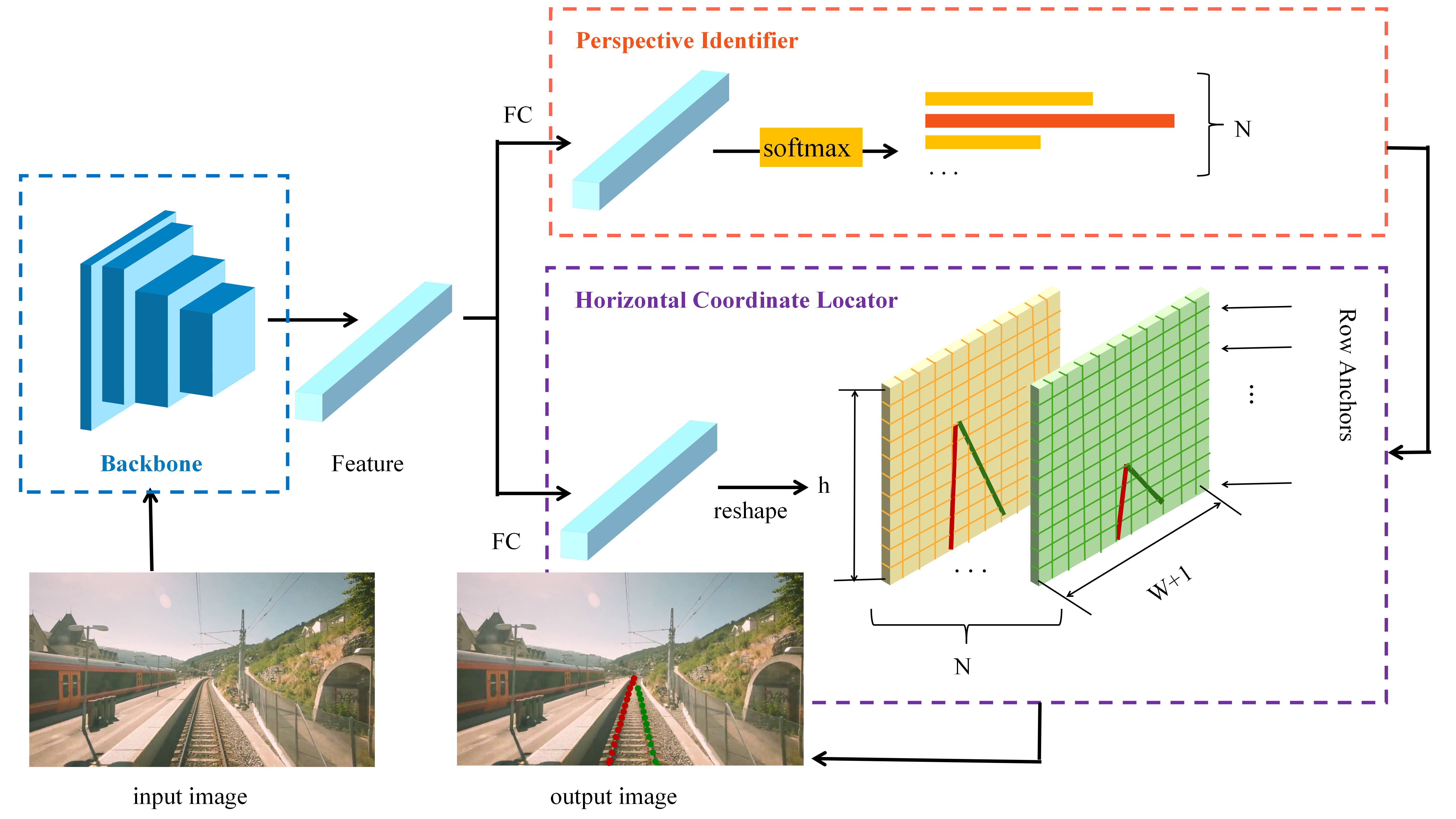}}

  \caption{The structure of UFATD. The image is first input into the backbone network to extract high-level features, and then it is fed into two separate branch networks. The \textit{Perspective Identifier} branch classifies the image category based on the camera perspective. Each category corresponds to a group of anchors. This branch is an  \(n\)-class classification network. The \textit{Horizontal Coordinate Locator} branch outputs \( ((w+1), h, C, n)\)-dimensional data, which includes the column coordinates under \(n\) groups of anchors. The classification result from the \textit{Perspective Identifier} branch is passed through a softmax function and then input to the \textit{Horizontal Coordinate Locator} branch, where the corresponding group of anchors for the image is selected.}
\label{fig:network}
\end{figure}

Ultra-fast adaptive track detection network (UFATD) is a row-selective track detection network.
Figure \ref{fig:network} illustrates the network architecture of UFATD.
This network leverages a pre-trained backbone to extract high-dimensional features from images. These features undergo two-dimensional convolution before being fed into two branch networks for dimensionality reduction of the feature map. 
The \textit{Horizontal Coordinate Locator} branch is tasked with classifying the row coordinates of the rails.

The classifier of the \textit{Horizontal Coordinate Locator} comprises layers of linear transformations ranging from 1800 to \( ((w+1) \times h\times C\times n)\) linear layers. Cross-entropy (CE) function is employed within the classification network to guide parameter optimization.
Assuming \(X\) represents the global features of the image, \(f^{ijk}\) is the classifier utilized to select the track position at the \(i\)-th track, the \(j\)-th row anchor, and the \(k\)-th set of the anchor groups. Then the railway track prediction can be written as:

\begin{equation}
    P_{i,j,k} = f^{ijk}(X), s.t. \quad i \in [0, C-1], j \in [0, h-1], k \in [0,n-1]
\end{equation}

among them, \(P_{i,j,k}\) is a \((w+1)\)-dimensional vector, indicating the probability of selecting (w+1) grid cells under the \(i\)-th railway track, \(j\)-th row anchor, and \(k\)-th set of anchor groups. Assume that \(T_{i,j,k}\) is a one-hot label at the correct location, the corresponding optimization formula is:
\begin{equation}
    L_{hcl} = \sum_{i=0}^{C-1}\sum_{j=0}^{h-1}\sum_{k=0}^{n-1}  L_{CE}( P_{i,j,k},T_{i,j,k})
\end{equation}
where \(L_{CE}\) is the cross entropy loss.

The \textit{Perspective Identifier} branch functions to identify the most suitable group of anchors from a variety of preset groups.
This branch serves as a classification network tasked with converting vectors into predictions of anchor group indices of size 
\(n\). The classifier within this branch comprises layers of linear transformations, spanning from 1800 to \(n\) linear layers.
After that, the results are processed through softmax, and the prediction corresponding to the image is selected from the results of the \textit{Horizontal Coordinate Locator}  branch.
Similar to the classification network mentioned earlier, the \textit{Perspective Identifier} branch also employs the CE function to guide parameter optimization during the training phase.
\begin{equation}
    L_{pi} = - \sum_{k=0}^{n-1} a_k \log(p_k)
\end{equation}
Each  \(p_k\)  represents the probability that the sample belongs to the \(k\)-th category, and the corresponding true label is a one-hot encoding vector \(\mathbf{a} = [a_0, a_1, \dots a_{n-1}] \), where \( a_k\) signifies whether the \(k\)-th category is the true category of the anchor groups. The full objective of the UFATD includes \(L_{hcl} \)and  \(L_{pi}\):

\begin{equation}
    L_{total} =  L_{hcl} + \lambda   L_{pi}
\end{equation}
where \(\lambda\) controls the \textit{Perspective Identifier}  branch. The specifics of the UFATD training process are delineated in Section \ref{sec:exp_set}.

\subsection{Anchor Genenration}
\label{sec:anchor_gen}

As illustrated in Figure \ref{fig:camera_angle}, the camera projection relationship affects the appearance of railway tracks in the image, resulting in the tracks appearing more curved at the far end of the camera, necessitating denser anchors for accurate representation. However, employing equidistant anchors would require a larger quantity of anchors, leading to an increase in network parameters.

In order to maximize the use of all anchors in the anchor group, an algorithm is proposed to generate non-equidistant anchor groups. 
Notably, the distance between anchor rows at the top of the image is closer compared to the bottom.

The points on the railway tracks that are closest to the top of the image are extracted, and their \(y\) values are recorded. The minimum \(y\) and maximum \(y\) values from \(y\) set are identified and denoted as \(y_{min}\) and \(y_{max}\), respectively.
The starting points of the \(n\) group of anchors are evenly distributed, so the starting value of the \(k\)-th group of anchors is calculated as follows:\(s_k = y_{min}+ \frac{k}{n-1}  \cdot (y_{max}-y_{min}) \)

Each group of anchors is then calculated based on its starting value and the maximum anchor value \(H\).
Specifically, the initial distance between anchor rows (i.e., equidistant anchors) is computed using the starting value, the maximum anchor value \(H\), and the number of anchors \(h\). In practice, the maximum anchor value will generally be smaller than image height \(H\). For simplicity, it is represented by \(H\).
The initial distance of the \(k\)-th group of anchors can be expressed as:
\(d_k = \frac{H - s_k}{h-1}\).
Subsequently, a scaling factor is calculated using Formula \ref{formula:1}, which is multiplied by the \(d_k\) to obtain the target distance. 
This calculation results in smaller growth for the first few points of each anchor group, leading to denser anchor placement and shorter distances between anchor rows near the top of the image.

\begin{equation}
f(x) = \begin{cases}
\sqrt{1 - (1 - x)^2} & \text{if } 0 \leq x \leq 1 \\
-\sqrt{1 - (1 - x)^2} + 2 & \text{otherwise}
\end{cases}
\label{formula:1}
\end{equation}

where \(x\) is defined as follows:
\(x_j =  \frac{j}{h} \cdot 2 \). 
And \(d_{j,k} = d_k \cdot f(x_j) \). Therefore, the \(j\)-th value in the \(k\)-th group of anchors is calculated as follows:
\begin{equation}
   y_{j,k} = s_k + \sum_{j=0}^{h-1} d_{j,k}  
\end{equation}
    
\section{Experiments}
\label{sec:exp}

\subsection{Implementations}
\label{sec:exp_set}
To quantitatively assess the proposed method, comprehensive experiments were carried out across three railway track datasets:  \textit{SRail}, \textit{RailDB} and \textit{DL-RAIL}.

\textbf{Dataset splits.}
The training datasets in the datasets randomly allocate 85\% of the images for training and 15\% for validation during the training phase. During each experiment, the training datasets and validation datasets are randomly partitioned.

\textbf{Training Details.}
All models are trained and tested using PyTorch on Nvidia RTX 4090 GPU, Intel i9 13900K CPU, with 64GB of RAM. 
The Adam optimizer is utilized, with a cosine decay learning rate strategy. The initial learning rate for the backbone is set to 4e-4. For the \textit{Perspective Identifier}, the initial learning rate is 5e-5, and for the \textit{Perspective Identifier}, it is 1e-3.
The batch size is set to 32, with a total training epoch number is 100.
Initially, the backbone network and the \textit{Perspective Identifier} branch are frozen. 
The pre-trained backbone network is utilized to first update the parameters of the \textit{Horizontal Coordinate Locator} branch. 
After 5 epochs, the backbone network is unfrozen. 
After 15 epochs, the \textit{Perspective Identifier} branch is unfrozen. Additionally, during the first 15 epochs, the \(\lambda\) controlling the \textit{Perspective Identifier} branch loss value is set to 0, and then it is set to 0.05.
The results of UFATD represent the average of three repeated experiments.

\textbf{Data Pre-process.} 
Furthermore, all images in the dataset will be resized to \(800\times288\) during training.
For the RailDB dataset, data processing has been introduced in Section \ref{sec:data_collect}.

\textbf{Evaluation Metric.}
All datasets utilize F1 as defined in CULane.
The calculation code comes from Qin \textit{et al} \footnote{https://github.com/cfzd/Ultra-Fast-Lane-Detection}.
\(FP\) represents false positives, and \(FN\)  represents false negatives, respectively. F1 is defined as follows:

\begin{equation}
    F1  = 2 \times \frac{Precision \times Recall}{Precision + Recall}
\end{equation}

where \( Precision = \frac{TP}{TP + FP}\), \(Recall = \frac{TP}{TP + FN}\).
In this study, the F1@50 and F1@75 metrics, with IoU thresholds of 0.5 and 0.75, are utilized. Additionally, to better compare the performance of the models, mF1 is also presented as follows: \(mF1 = (F1@50 + F1@55 + \cdots +F1@95)/10\)

\subsection{Ablation Study}

Performing an ablation analysis on UFATD  on SRail to validate the effects of the backbone network, the number of anchors, and the inter-anchor distance on the network. Except for the above conditions, all experiments were conducted following the identical protocols outlined in Section \ref{sec:exp_set}.

\subsubsection{Backbone of UFATD}
Beyond the evaluation of ResNet18, ResNet34 and ResNet50 \citep{he2016deep} as backbone architectures, ResNet 34 with Fcanet \citep{qin2021fcanet}, SqueezeNet \citep{iandola2016squeezenet} and MobileNet V2 \citep{sandler2018mobilenetv2}, were also investigated to elucidate the influence of UFATD across a spectrum of backbone architectures. Table \ref{tab:backbone_results} outlines the F1 score and Frames Per Second (FPS) outcomes for different backbone configurations.

\begin{table}[ht]
\centering
\caption{Quantitative comparison of various backbones on the SRail dataset.}
\label{tab:backbone_results}
  \begin{tabular}{ l  c  c c c c c}
  
    \toprule
Backbone     &Res18  &Res50 & Res34  & Res34 with Fcanet &SqueezeNet  &MobileNet V2 \\
\midrule
F1@50 & 98.68& 98.64&  \textbf{98.79}& 98.77&98.19& 98.55 \\
 FPS & 473 & 249 & 301&  231&  \textbf{951}&  732\\  

 \bottomrule
  \end{tabular}
\end{table}

The results demonstrate that ResNet models are relatively dominant, indicating that even a small network like ResNet18 is sufficient for railway feature extraction.
In addition, what is interesting is that under repeated experiments, the measured FPS of Resnet50 is indeed higher than that of Resnet34, which is incredible. When tested on Rtx4080 laptop, the FPS of Resnet34 is higher. This may involve the optimization of the graphics card, which is worth studying.
In repeated tests, Resnet50 consistently outperforms Resnet34 on FPS, which is remarkable. However, when tested on an RTX4080 laptop, Resnet34 achieves higher FPS. This discrepancy suggests potential optimizations related to the GPU, prompting further investigation.

\subsubsection{Effectiveness of the proposed modules}
This section demonstrates the impact of the new network proposed in Section \ref{sec:network} and the anchors generated by the algorithm introduced in Section \ref{sec:anchor_gen} on model performance.
As observed in Table \ref{tab:eq_non}, the network proposed in this study notably enhances the F1@75 and mF1 metrics of the model. Nonetheless, there are some drawbacks. 
F1@50 decreased by 0.44, likely attributed to the addition of multiple groups of anchors, thereby complicating the classification process for the  \textit{Horizontal Coordinate Locator} branch. 
Moreover, employing non-equidistant anchors generated via the proposed anchor generation method also contributes to a moderate improvement in the model's performance.

\begin{table}[ht]
\centering
\caption{Experiments of the proposed modules on SRail with ResNet-18 backbone. The baseline represents training the network using only a group of equidistant anchors.
The new network introduces three groups of equidistant anchors.
Anchor generation involves using three groups of non-equidistant anchors, produced by the anchor generation algorithm, to train the network.}
\label{tab:eq_non}
\begin{adjustbox}{width=1.\textwidth} 
  \begin{tabular}{ c  c  c l  l  l}
  
    \toprule
Baseline &  New network  & Anchors generation & mF1 & F1@50 & F1@75   \\  
\midrule
\checkmark & & & 54.60& 98.94 & 56.80 \\
 & \checkmark& & 57.06 (+2.46)&  98.50 (-0.44)& 65.05 (+8.25)\\
  & \checkmark & \checkmark& 58.12 (+3.52)& 98.68 (-0.26)& 67.68 (+10.88)\\
 \bottomrule
  \end{tabular}
  \end{adjustbox}
\end{table}

\section{Results and Discussion}
In this section, the outcomes of UFATD and several models on the railway track dataset will be showcased. ResNet-18 and ResNet-34 were employed as the backbone architectures for UFATD.

\begin{table}[b]
\centering
\caption{Quantitative comparison of various methods on the SRail dataset.}
\label{tab:SRail_results}
\begin{adjustbox}{width=1.\textwidth} 
  \begin{tabular}{ l | c  c c  c c  c c  c c}
  
   \toprule
    Method  & mF1 &  F1@50 &  F1@75 & FPS  & Class 0& Class 1 & Class 2 & Class 3 \\ 
    \midrule

    Be´zierLaneNet \citep{feng2022rethinking} & 43.77 & 90.38 & 36.28& 364 &84.01 &85.58& 93.20&92.84 \\
    Clrnet \citep{zheng2022clrnet} & \textbf{64.03} &98.62	&\textbf{79.55}&279	&97.30	&99.06	&99.14	&97.56	\\
    CondlaneNet \citep{liu2021condlanenet} &	56.17 &\textbf{99.31}	&55.96 & 327	&\textbf{99.53}	&\textbf{99.61}	&\textbf{99.46}	&\textbf{98.13} \\
    DALNet \citep{yu2024dalnet} & 60.40 & 98.52& 68.59 & 305& 96.28& 99.06& 99.35& 97.15\\
    Rail-Net \citep{zheng2021rail} & 51.24 & 97.97 & 46.97 & 429& 97.0 & 97.19 & 98.70 & 97.27\\ 
    UFLD (Res18) \citep{qin2020ultra}  & 49.81&  97.71 & 42.49 & \textbf{475}   &96.28 & 96.25 & 98.76 & 97.25 \\ 
    UFLD (Res34)  \citep{qin2020ultra} &49.58 &97.95& 40.90 & 249 & 96.10 &96.72 & 98.92 &97.94\\
    
   \midrule
    UFATD (Res18)  & 58.12 & 98.68	& 67.68 & 473 & 98.88 & 98.65 & 98.97& 97.46\\ 
    UFATD (Res34) & 58.04 & 98.64& 	67.59	&  249& 98.47& 	98.48& 	98.79	& 97.17	\\
    UFATD (Res34 with Fcanet)&	58.19 &98.78	&68.23&	231&98.57&	99.01&	99.10&	97.53\\

    UFATD (Res50) & 57.92 & 98.80	&67.35	& 301 &98.76&	98.75	&99.08	&97.85	\\

    \bottomrule
  \end{tabular}
  \end{adjustbox}
\end{table}

On the S-Rail dataset, UFATD was compared with six methods, including: Be´zierLaneNet \citep{feng2022rethinking}, Clrnet \citep{zheng2022clrnet},  CondlaneNet \citep{liu2021condlanenet}, DALNet \citep{yu2024dalnet}, Rail-Net \citep{zheng2021rail}, UFLD \citep{qin2020ultra}.
This experiment conducted a comparative analysis of model accuracy and execution time. Model accuracy was assessed using the F1 score metric, while model runtime was measured as the average time over 1000 iterations. 
From Table \ref{tab:SRail_results}, it can be inferred that UFATD demonstrates performance on par with SOTA while exhibiting high operational speed.

On RailDB, UFATD is SOTA in multiple categories. It is worth noting that, even in cases where models such as UFLD perform poorly, UFATD still demonstrates reliability. Refer to Table \ref{tab:RailDB_results}.

\begin{table}
\centering
\caption{Quantitative comparison of various methods on the RailDB dataset.}
\label{tab:RailDB_results}
\begin{adjustbox}{width=1.\textwidth} 
  \begin{tabular}{ l | c  c c  c c  c c  c c c c c c}
  
    \toprule
Method  & mF1 & F1@50 & F1@75   & Cross  &  Curve  & Far  & Line &  Near  & Night &  Rain  & Slope  & Sun \\ 

\midrule
Be´zierLaneNet \citep{feng2022rethinking} &42.30 &	82.75 &	38.07 &	48.34 &	91.92	 &93.99	 &86.48	 &79.85	 &78.54 &	83.22	 &92.95 &	82.24\\
Clrnet \citep{zheng2022clrnet} &40.25 &	82.00 &	33.45	 &70.03	 &78.04 &	75.20 &	97.30	 &89.14	 &78.70 &	82.50 &	92.27	 &81.24\\
 CondLaneNet \citep{liu2021condlanenet}&55.10	&90.95	&61.33	&56.80	&88.11&	91.20	&86.24&	79.70	&92.20&	95.73	&96.58&	83.56\\
UFLD (Res18) \citep{qin2020ultra}&  54.41&	95.52&	57.60	&82.23	&97.99&	92.79	&95.82	&98.71&	97.45	&94.51	&96.94&	96.19\\
Rail-Net  \citep{zheng2021rail}& 52.64	& 97.02	& 51.12	& 89.95	& 97.99	& 94.90& 	97.41	& 99.11& 	98.72	& 95.93	& 98.34	& 95.97 \\ 
GANet \citep{wang2022keypoint}&58.87	&91.59	&68.40	&53.37&	98.34	&\textbf{95.95}	&\textbf{98.46}&	90.16&	93.24	&\textbf{96.07}&	97.01&	84.55\\
\midrule
UFATD (Res18)  & 59.97& 	97.43& 	67.70	& 91.54	& \textbf{98.41}& 	95.25& 	97.96& 	99.12	& 98.13	& 95.85	& 98.87	& \textbf{96.47} \\ 
UFATD (Res34)  &\textbf{60.53}& 	\textbf{97.60}	& \textbf{68.90}& 	\textbf{92.10}& 	98.37	& 95.60	& 98.13& \textbf{99.24}	& \textbf{98.70}	& 96.03	& \textbf{98.92}	& \textbf{96.47}\\

 \bottomrule
  \end{tabular}
   \end{adjustbox}
\end{table}

\begin{figure}
  \centering
  {\includegraphics[width=0.7\textwidth]{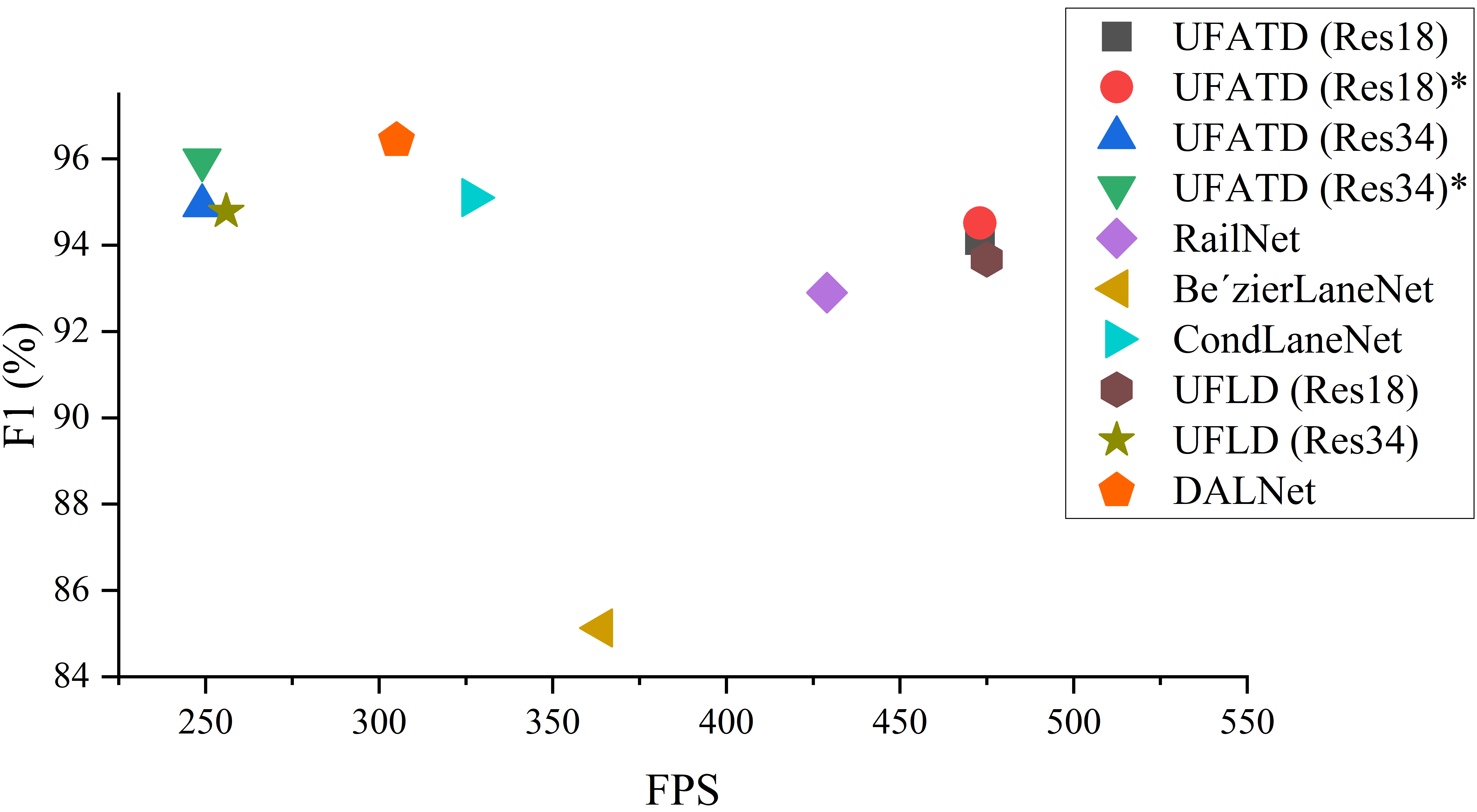}}

  \caption{Quantitative comparison of various methods on the DL-RAIL dataset. \(^*\) indicates that the test set is employed as the validation set for the optimal model.}
\label{fig:DL-Rail_results}
\end{figure}
\begin{figure}
  \centering
  \subfigure[SRail]{\includegraphics[width=0.325\textwidth]{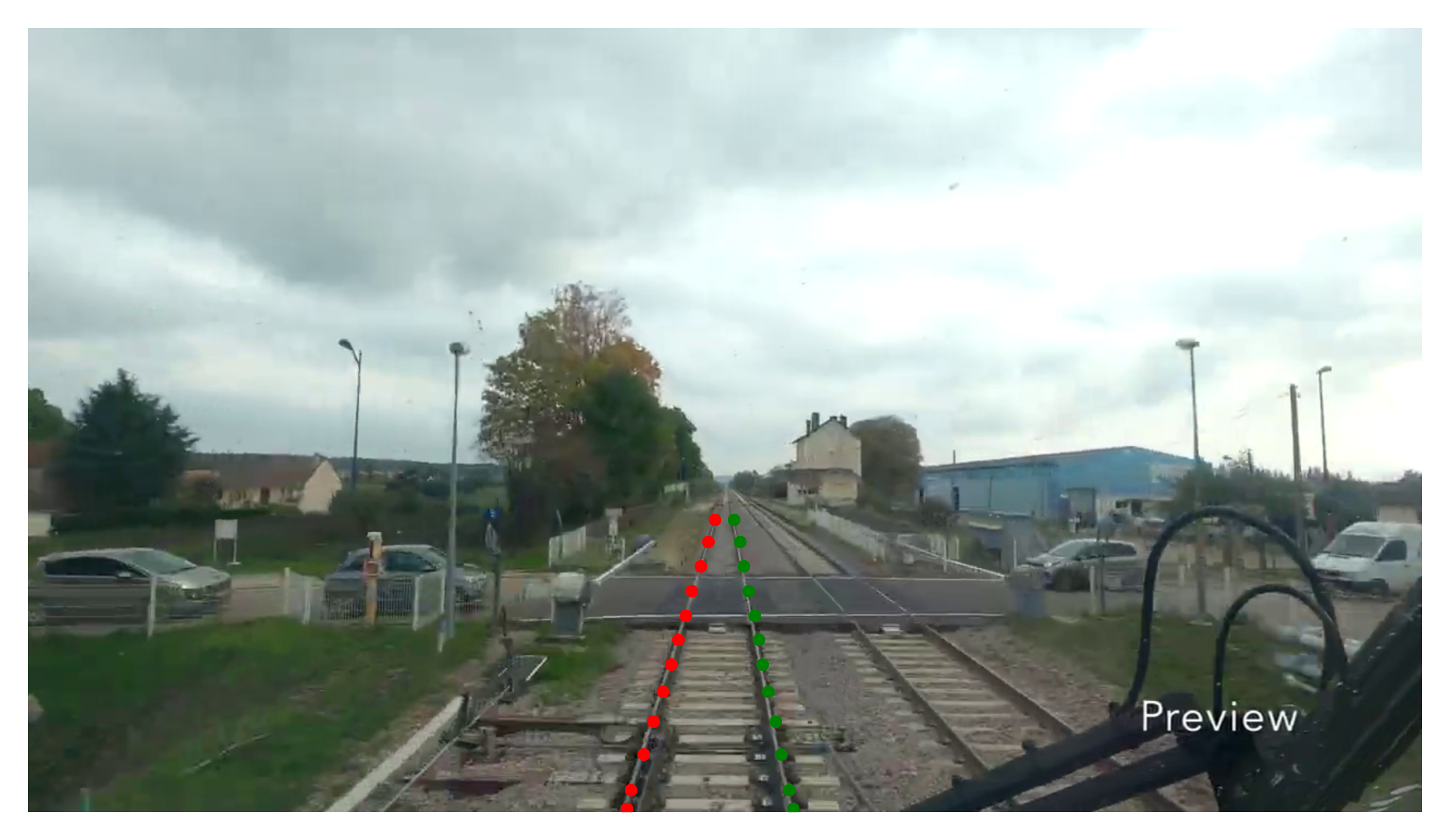}}
  \subfigure[Rail-DB]{\includegraphics[width=0.325\textwidth]{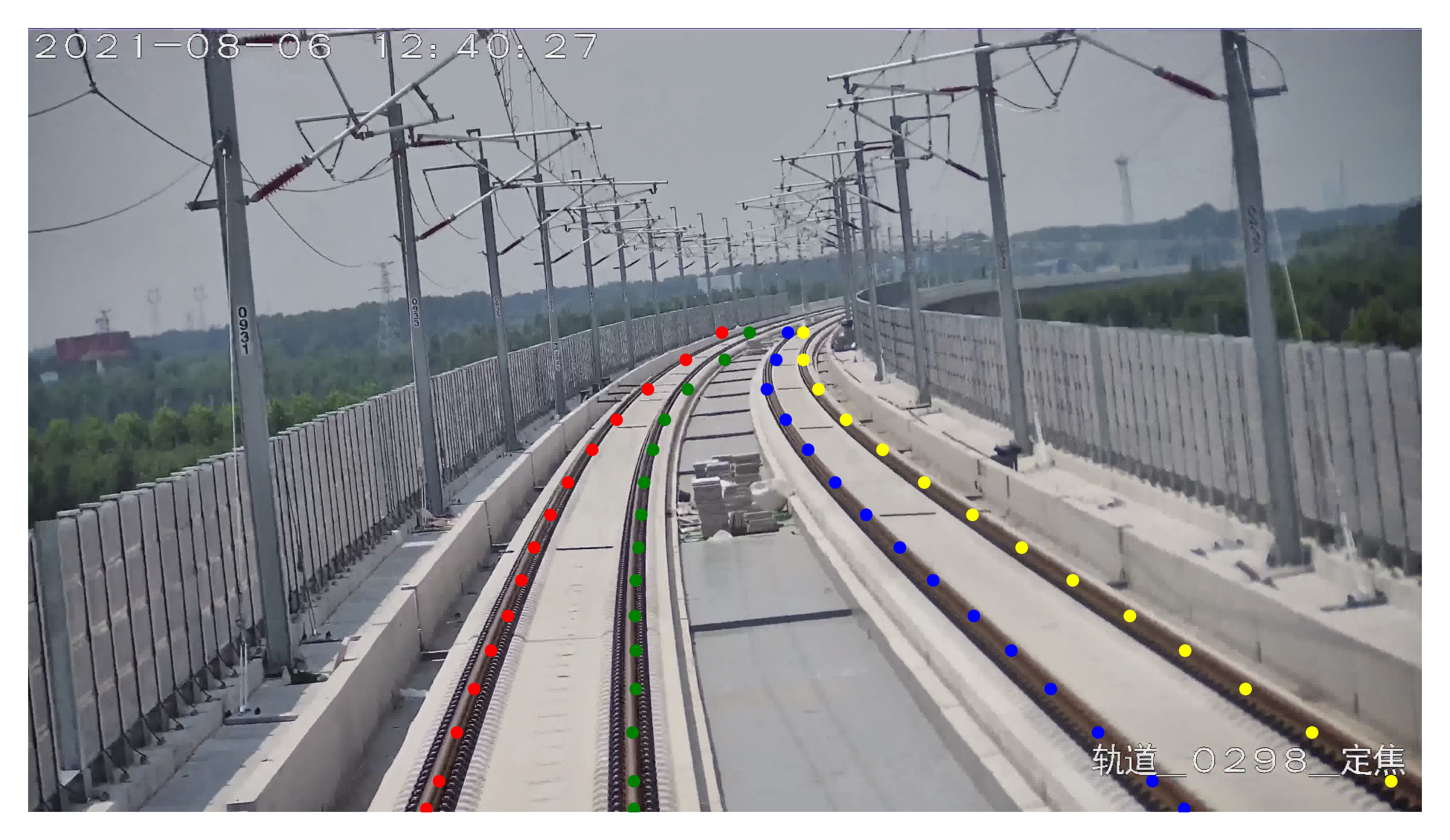}}
  \subfigure[DL-RAIL]{\includegraphics[width=0.325\textwidth]{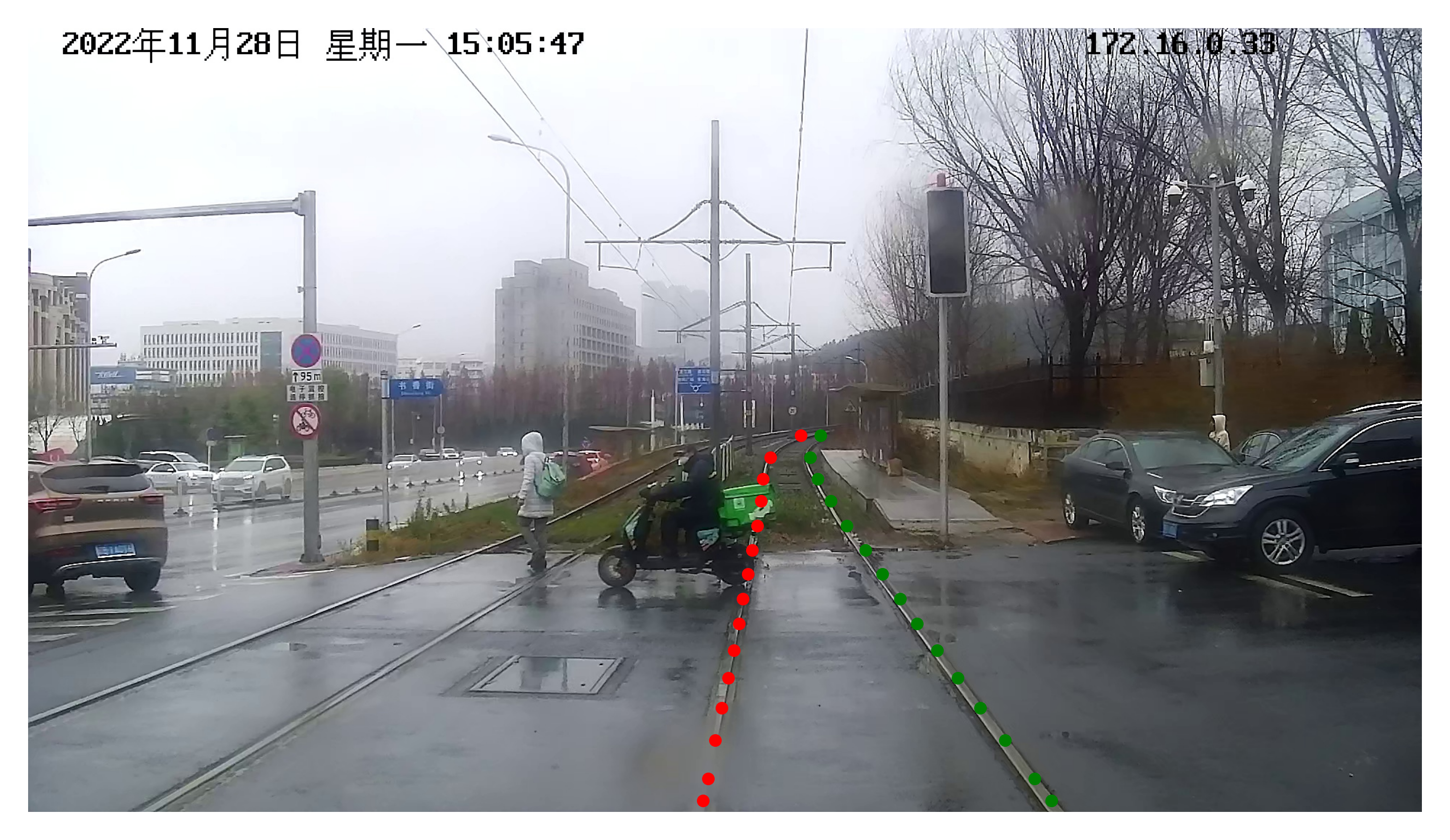}}

  \caption{Visualization on the SRail, the Rail-DB and the DL-RAIL dataset. }
\label{fig:vis}
\end{figure}

For the DL-Rail dataset, UFATD demonstrates outstanding performance as well. In the code provided by Yu \textit{et al.} \citep{yu2024dalnet}, the test set is involved in the validation process during model training. 
Consequently, Figure \ref{fig:DL-Rail_results} provides two types of results. Those marked with an asterisk (*) represent model outcomes when the test set is employed as the validation set. UFATD with ResNet-34 achieves a performance ranking second only to DALNet.
The visualization of UFATD is depicted in Figure \ref{fig:vis}. UFATD exhibits strong performance across all three datasets.

\section{CONCLUSION}
\label{sec:con}

This paper introduces an adaptive railway track detection model capable of accommodating various camera perspectives and achieving outstanding levels of speed and accuracy. The proposed model addresses track detection by framing it as a row-based selection and anchor category classification challenge.
The experiments demonstrate that the lightweight version of UFATD achieves a speed of 473 FPS while maintaining exceptionally high accuracy.

In the future, our emphasis will be directed towards addressing the following challenges: exploring methodologies such as feature fusion to enhance the model's feature extraction capabilities and minimize the number of anchors.
Additionally, this work can also be applied to lane line detection, but more research is needed.

\begin{ack}

Funding in direct support of this work: Natural Science Foundation of Hunan Province, China grant No.S2024JJMSXM1606, Hunan Provincial High
tech Industry Science and Technology Innovation Leading Plan Grant No.2020GK2096. And this work was supported in part by the High Performance Computing Center of Central South University.

\end{ack}

\bibliographystyle{plainnat}
\bibliography{references}

\medskip





\end{document}